\documentclass[submission,copyright,creativecommons]{eptcs}
\providecommand{\event}{ICLP 2020} 

\usepackage{breakurl}             
\usepackage{underscore}           

\usepackage{mathptmx}

\usepackage{url}  
\usepackage{graphicx}  
\usepackage{amsmath}
\usepackage{amssymb}
\usepackage{amsfonts}
\usepackage{listings}

\usepackage{amsthm}

\title{An application of Answer Set Programming\\ in Distributed Architectures: ASP Microservices}

\author{Stefania Costantini
\institute{Dipartimento di Ingegneria e Scienze dell'Informazione e Matematica\\
Universit{\`{a}} degli Studi di L'Aquila, L'Aquila, Italy}
\email{stefania.costantini@univaq.it}
\and
Lorenzo De Lauretis
\institute{Dipartimento di Ingegneria e Scienze dell'Informazione e Matematica\\
Universit{\`{a}} degli Studi di L'Aquila, L'Aquila, Italy}
\email{\quad lorenzo.delauretis@graduate.univaq.it}
}

\def\titlerunning{Microservices in Answer Set Programming}
\def\authorrunning{S. Costantini \& L. De Lauretis}
\begin{document}
\maketitle

\newcommand{\ind}{\hspace{1cm}}
\newcommand{\st}{\medskip\noindent}
\newcommand{\stn}{\medskip}
\newcommand{\noi}{\noindent}
\def\ni{\noindent}
\newcommand{\sni}{\smallskip\noindent}
\newcommand{\bs}{\bigskip}
\newcommand{\sms}{\smallskip}
\newcommand{\ms}{\medskip}
\newcommand{\tbs}{\hspace*{4mm}}
\newcommand{\tbm}{\hspace*{8mm}}
\newcommand{\tbl}{\hspace*{12mm}}
\def\naf{not\,}
\def\ACC{\mathit{ACC}}

\newcommand{\G}{\Gamma}
\newcommand{\hb}{\rm I \!B}
\newcommand{\es}{\emptyset}
\newcommand{\lt}{\left\{}
\newcommand{\rt}{\right\}}
\def\and{ \ \wedge}
\newcommand{\ut}{ \ \wedge}
\def\implies{ \supset}
\newcommand{\sm}{\setminus}
\newcommand{\ifandif}{\;\Leftrightarrow\;}
\def\ar{\leftarrow}
\newcommand{\rar}{\rightarrow}
\newcommand{\then}{\Rightarrow}
\newcommand{\onlyif}{\Leftarrow}
\newcommand{\eqdef}{\stackrel{\rm def}{=}}
\newcommand{\nin}{\not\in}
\newcommand{\hyp}{\stackrel{\models}}
\def\ss{\supset}
\newcommand{\om}{\omega}
\newcommand{\bars}{\!\! \parallel \!\!}
\newcommand{\nuP}{\nu^{\prime}}

\newcommand{\prol}{:\!\!-\,} 
\newcommand{\bfnot}{{\bf not}}
\newcommand{\notexists}{\exists\mkern-7mu /}
\newcommand{\nonmdash}{\mid\mkern-4mu\sim}
\newcommand{\no}{\hbox{\it not}\ }

\newcommand{\ov}{\overline}
\newcommand{\un}{\underline}
\newcommand{\li}{\mbox{\textbf{---\ \,}}}
\newcommand{\lili}{\mbox{\textbf{-----\ \,}}}
\newcommand{\as}{``}
\newcommand{\ad}{''}
\newcommand{\be}{\begin{em}}
	\newcommand{\ee}{\end{em}}
\newcommand{\bb}{\begin{bf}}
	\newcommand{\eb}{\end{bf}}
\newcommand{\I}[1]{\relax\ifmmode\mbox{\it#1}\else{\it#1}\fi}
\newcommand{\meno}{\medskip\noindent}
\newcommand{\sno}{\smallskip\noindent}
\newcommand{\rif}{~\ref}
\newcommand{\PROL}{\mbox{:-}}
\newcommand{\IF}{\leftarrow}

\newcommand{\caln}{\mathcal{N}}
\newcommand{\cali}{\mathcal{I}}
\newcommand{\calr}{\mathcal{R}}
\newcommand{\calt}{\mathcal {T}}
\newcommand{\calv}{\mathcal {V}}
\newcommand{\cals}{\mathcal {S}}
\newcommand{\calh}{\mathcal {H}}
\newcommand{\calm}{\mathcal {M}}

\newcommand{\K}{\mbox{\textbf{K}}}
\newcommand{\M}{\mbox{\textbf{M}}}
\newcommand{\N}{\mbox{\textbf{not}}\,}
\newcommand{\NO}{\mbox{\textbf{NOT}}\,}
\newcommand{\Kw}{\mbox{\textbf{K}$^W$}}
\newcommand{\Mw}{\mbox{\textbf{M}$^W$}}
\newcommand{\Nw}{\mbox{\textbf{\no}$^W$}}
\newcommand{\NOw}{\mbox{\textbf{NOT}$^W$}\,}
\newcommand{\Mu}{\mbox{$\mu$ASPSv}}
\newcommand{\Mus}{\mbox{$\mu$ASPSv's}}

\newcommand{\calmP}{\calm^{\prime}}
\newcommand{\caliP}{\cali^{\prime}}
\newcommand{\bbbb}{{\rm I\!B}}

\newcommand{\Pip}{{\Pi}^{\prime}}

\newtheorem{definition}{Definition}
\newtheorem{lemma}{Lemma}[section]

\begin{abstract}
We propose an approach to the definition of
microservices with an Answer Set Programming (ASP) `core',
where microservices are a successful abstraction for 
designing distributed applications as suites of independently deployable
interacting components. 
Such ASP-based components might be employed
in distributed architectures related to Cloud Computing or
to the Internet of Things (IoT).
\end{abstract}

\section{Introduction}

An important hot topic is, in our opinion, that of defining software engineering
principles for Answer Set Programming (ASP). ASP is a successfully logic 
programming paradigm (cf. \cite{ASPJournal2016} and the references therein) stemming from the Answer Set (or \as Stable Model'') Semantics of Gelfond and Lifschitz, and based on the programming methodology proposed by
Marek, Truszczy{\'{n}}ski and Lifschitz. ASP is put into practice by means of effective inference engines, called \emph{solvers}\footnote{Many performant ASP solvers are freely available, a list of them is reported at \url{https://en.wikipedia.org/wiki/Answer_set_programming}.}.
ASP has been widely applied, e.g., to information integration, constraint satisfaction, routing, planning, diagnosis, configuration, computer-aided verification, biology/biomedicine, knowledge management, etc. .

It is important in our view to have the possibility of defining significant subprograms as \emph{components}
to be possibly distributed over different nodes of a network. Both the connections between components and the ways of exchanging information should be clearly defined.
Our approach is inspired by the \emph{microservices} architectural abstraction, which can be described as a particular way of designing distribute software applications as suites of independently deployable interacting services (cf. for instance the survey \cite{JamshidiPMLT18} or see 
{\small\url{https://martinfowler.com/articles/microservices.html#CharacteristicsOfAMicroserviceArchitecture}}). A microservice is indeed a component, as it is a unit of software that is independently replaceable and modifiable: in fact, it intended as a self-contained piece of business functionality with clear interfaces that can be accessed by the \as external world''. This kind of architectural abstraction enables distribution, as each microservice is meant to be executed as an independent process, and heterogeneity, allowing for different services to be written in different programming languages. Microservices are a suitable architectural abstraction for the Internet of Things (IoT): a microservice may incapsulate a physical objects, where service inputs and/or outputs can possibly be linked to sensors/actuators. Since microservices are by their very nature heterogeneous, open issues are: the way microservices communicate with each other (synchronous, asynchronous, which is the message format, etc.); and, the protocols used for the communication.

Microservices in real distributed software architectures and in cloud computing are usually deployed via \emph{lightweight containers}. A container in software engineering is a standard unit of software that packages up code and all its dependencies so the application runs quickly and reliably and can be seamlessly transferred from one computing environment to another. A widely used tool to
create containers is \emph{Docker}, available in the form of an open source Doker Engine (see \url{https://www.doker.com}). A Docker container image consists in a lightweight, standalone, executable package of software that includes all elements needed to run an application: code, runtime support, system tools, system libraries and settings. 

Along this line, we propose ASP microservices that might be blended into heterogeneous systems, and even into Multi-Agent-System (MAS) since each such component may be seen as a reactive agent. They could in perspective be employed in cloud computing, and in IoT, including robotic applications.
In this paper we discuss a first idea about how these components, that we call $\mu$ASP-Services ($\mu$ASPSv's), can be specified, how their interfaces to the \as external world'' can be defined, and how they should procedurally behave. In fact, a $\mu$ASPSv is meant to be based upon a `core' ASP program whose activities, however, should be triggered by external stimula/requests coming from some source, and whose results should be returned to the requesters.
Therefore, the `core' ASP program should be included into a container, that can be possibly realized via the Docker technology, which should also include: an interface, to provide the \Mu\ with inputs, and to deliver the outputs; solving capabilities to compute the answer sets. For ASP, standalone versions of the most important solvers are nowadays available. So, a docker deployment for a \Mu\ should include the source program, its `execution shell', and the solver. 
 
A small specimen of the proposed approach is represented in
the following example, which is meant to be (a fragment of) the
code of a \textit{controller} component/agent, acting in the IoT. This piece of code might be in fact the ASP `core' of a \Mu.
$test\_ok$ is the input 
coming from a sensor, with value `true' if the controlled device is working properly,
(otherwise the value is set to false).

\smallskip
$
\begin{array}{l}
\mathit{device\_ok}\, \ar\ \mathit{test\_ok}.\\
\mathit{device\_fault}\, \ar\ \no\, \mathit{test\_ok}.\\
\mathit{wait} \ar\, \no\, \mathit{wait},\ \no\, \mathit{sensor\_input}.
\end{array} $

\smallskip
In this example, as a programming choice,
inconsistency (due to the odd cycle over $wait$) is to be
interpreted as a `no-operation' controller state, where the component is waiting for
the sensor outcome. 
It can be assumed that the sensor provides results at a 
certain frequency. The outcome, i.e., $\mathit{device\_ok}$ or $\mathit{device\_fault}$, is to be delivered
to whatever other components would ask for it.

More precisely, in order to work in a standalone way within a distributed system, an interface (or `shell') will manage the ASP program, and in particular will perform the following functions. First, manage the inputs and outputs of the \Mu: i.e., be able to detect input arrival and to dispatch the outputs according to the request coming from the \Mu's external environment. In the above example, inputs can be: (1) queries over the device state for which an answer has to be delivered, and (2) sensor outcomes, which are to be considered as particular inputs which activate the module. In the general case, upon the arrival of inputs, the shell will: (i) add the inputs to the ASP program as facts; (ii) evaluate the answer sets of the resulting ASP program; (iii)
according to previously-received requests, extract (from the answer sets) the answers and and deliver them to the external environment. Notice that the shell, after delivering the outputs, will remove the last-added program facts so as to bring back the controller to the `no-operation' state. In a `stateless component', all inputs will be removed, while some of the inputs can be left if instead the component is meant to have a state.

In this paper we introduce a formal definition of ASP microservices and we outline a possible logic-based semantics of an overall heterogeneous distributed system encompassing such modules. 
The paper is in fact structured as follows. In Section\rif{backmicro} we introduce basic concepts about microservices. In Section\rif{asp} we recall (for the sake of completeness) the Answer Set Programming paradigm, and in Section\rif{aspmod} we briefly survey and discuss existing approaches to modularity in ASP. We introduce {\bf our contribution} in Sections\rif{muspecification} and\rif{extendedMCS}, i.e.: (1) how to define and implement \Mus\ so as to be able to get inputs and extract answers, and how the inner ASP program might be structured; (2) how to provide a formal semantics to a generic microservice architecture possibly encompassing \Mus. In Section\rif{casestudy} we discuss a small case study, developing a specific \Mu\ which implements an intelligent agent managing a road intersection (i.e., a \as virtual traffic light''). Finally, in Section\rif{conclusions} we conclude.

\section{Background: Microservices}
\label{backmicro}

To better understand Microservices, let us introduce before the concept of \as Service''. A Service, as a software component, is a mechanism to enable access to one or more software capabilities~\cite{service:2015}. It provides other applications with stable, reusable software functionality at
an application-oriented, business-related level of granularity using
certain standards~\cite{service:2007}.
Service-Oriented Architecture (SOA) is a software architectural style that
uses services as the main building component~\cite{service:2015}.
Key features of SOA are heterogeneity, standardization and evolvability
of services.

Microservices can be seen as a technique for developing software applications
that, inheriting all the principles and concepts from the SOA style, permit to structure a service-based application
as a collection of very small loosely coupled software services~\cite{dragoni:2017}.

A MicroServices Architecture (MSA), is an evolution of the SOA architecture, making the communication lighter and the software parts (Microservices) smaller. It can be seen as a new paradigm
for programming applications by means of the composition of small
services, each running its own processes and communicating via light-weight
mechanisms. Key features of MSA are bounded
scope, flexibility and modularity~\cite{de2019monolithic}. I.e., there is a clear definition of the data a microservice service is responsible for and is \as bound to.'' So, the service owns this data and is responsible for its integrity and mutability.

The work in \cite{krivic2017microservices} shows that a distributed MSA can easily fit into an IoT system. In particular, the set of microservices can be seen as a Multi-Agent-System, cooperating to realize all system functionalities.

At the current day, microservices are still a new and emerging paradigm, having building standards not perfectly defined and communication protocols that are not well specified: in fact, following one of the definitions of microservices \cite{dragoni:2017,de2019monolithic}, they are small loosely coupled software services that communicate, possibly exploiting service discovery to find the route of communication between any two of them. 
In our work, we are proposing a new approach, that is $\mu$ASP-Services, which are based upon an inner ASP program. 

\section{Answer Set semantics (AS) and Answer Set Programming (ASP)}
\label{asp}

\as Answer Set Programming'' (ASP) (cf. \cite{ASPJournal2016} and the references therein) is a successful programming paradigm based on the Answer Set Semantics. In ASP, one can see an answer set program (for short, just \as program'') as a set of
statements that specify a problem, where each answer set
represents a solution compatible with this specification. Whenever a program has no answer sets (no solution can be found), it is said to be \emph{inconsistent}, otherwise it is said to be \emph{consistent}. 
 
Syntactically, an ASP program $\Pi$ is a
collection of \emph{rules} of the form

\(H\leftarrow\; A_{1} , \ldots , A_m , \naf\,A_{m+1}, \ldots, \naf\,A_{m+n}.\)

where $H$ is an atom, $m,n\geqslant 0$, and each $A_i$, $i \leq m+n$, is
an atom. Atoms and their negations are called \emph{literals}. Symbol $\leftarrow$ is often indicated as \PROL\ in practical programming.
The left-hand side and the right-hand side of the clause are called
\emph{head} and \emph{body}, respectively. A rule with empty body is called a \emph{fact}. 
A rule with empty head is
a \emph{constraint}, where a constraint of the form
`\(\leftarrow L_1,...,L_n.\)'
states that literals $L_1,\ldots,L_n$ cannot be simultaneously true
in any answer set. Constraints are often rephrased as `\(f\leftarrow \naf f, L_1,...,L_n.\)'
where $f$ is a fresh atom. To avoid the contradiction over $f$, some of the $L_i$'s must
be false thus forcing $f$ to be false, and this, if achieved, fulfills the constraint.
There are other features that for the sake of simplicity we do not consider in this paper. 

The answer set (or \as stable model'') semantics (AS) can be defined in several ways
(cf., e.g., \cite{Lifschitz08}, though more recently several other definitions have appeared in the literature). However, answer sets of a program $\Pi$, if any exists, are supported minimal classical models of the program interpreted as a first-order theory in the obvious way. 

The ASP approach to problem-solving consists basically in the following: (i) encoding of the given problem via an ASP program; (ii) computing the \as answer sets'' of the ground program via an ASP solver (as a preliminary step, solvers perform the \as grounding'' of the program, by substituting all variables with the constants occurring in the program); (iii) extracting the problem solutions by examining such answer sets; in fact, answer set contents can be in general reformulated in order to present the solution in terms of the given problem. 

A top-down (prolog-style) query answering device has been defined in \cite{CostantiniF13} for RAS, where RAS is a variation
of AS where every program admits answer sets. However, RAS and AS coincide over a wide class
of programs (some sufficient conditions that identify classes of programs where the two semantics
coincide are reported in \cite{Costantini19}). Queries that have been introduced are, first of all, \as ?\,$A$'' asking whether $A$ is true w.r.t. some answer set of given program $\Pi$. Other queries are the following:
query \as ?\,\N $A$'' asks whether $A$ is false w.r.t. some answer set of $\Pi$, and therefore it succeeds if $\no A$ is true in some of them (this implements the operator \N introduced in \cite{ShenE16});
query \as ?\,$\N \no A$'' asks whether $\no A$ is false in some answer set, and therefore it succeeds if $A$ is true in some of them, which corresponds to query \as ?\,$\M A$'', $\M$ standing for `possibility' in the modal logic sense;
query \as ?\,$\no$ \N $A$'' asks whether it is not true that $A$ is false w.r.t. some answer set of $\Pi$, i.e., that $A$ is true in all of them, which corresponds to \as ?\,$\K A$'', $\K$ standing for `knowledge' in the modal logic sense; 
query \as ?\,$\no \N \no A$'' asks whether $A$ is false in every answer set, meaning $\K\, \no A$, i.e., $\no \M A$ (a new operator \NO is a shorthand for $\no \N \no A$).
\section{Background: Modularity in ASP}
\label{aspmod}

Existing approaches to modularization of ASP programs have been reviewed in \cite{DyoubCG18}, to which the reader may refer for a complete account. Reporting from there, such approaches can be divided into two lines: \as programming-in-the-large'', where programs are understood as combinations of separate and independent components, combined by means of compositional operators; \as programming-in-the-small'', in which logic programming is enriched with new logical connectives for managing subprograms.

Considering the programming-in-the-small vision:
in \cite{eiter1997modular}, program modules are viewed as generalized quantifiers;
\cite{ianni2004enhancing} proposes templates for defining subprograms;
\cite{tari2005language} developed a declarative language for modular ASP, which allows a programmer to declaratively state how one ASP module can import processed answer sets from another ASP module. The work in \cite{Cos06} explores how to divide an ASP program into components according to its structure in terms of cycles.

Lifschitz and Turner's \as splitting set theorem'' (cf. \cite{lifschitz1994splitting}),
or variants of it, is underlying many programming-in-the-large approaches. The basic idea is that a program can be divided into two parts: a \as bottom'' part and a ``top'' part, such that the former does not refer to predicates defined in the latter. Computation of the answer sets of a program can be simplified when the program is split into such parts.

\cite{janhunen2009modularity} defines the notion of a \as DLP-function'' which is basically a module for which a well-defined input/output interface is provided; a suitable compositional semantics for modules is introduced. 
\cite{oikarinen2008modularity} provides a simple and intuitive notion of a logic programming module that interacts through an input/output interface. This is achieved by accommodating modules as proposed by \cite{gaifman1989fully} to the context of Answer Set Programming. Full compatibility of the module system with the stable model semantics is achieved by allowing positive recursion to occur inside modules only. 

\cite{dao2009modular} focuses on modular non-monotonic logic programs (MLP) under the answer set semantics, where modules may provide input to other modules. Mutually recursive module calls are allowed. 

\cite{baral2006macros} defines modules in terms of macros that can be called from a program. \cite{balduccini2007modules} provides modules specification with information hiding, where modules exchange information with a global state. 

In \cite{faber2009manifold} a technique is proposed to allow an answer set program to access the brave or cautious consequences of another answer set program. 
\cite{lierler2013modular} proposes \as modular logic programs'' as a modular version of ASP. This work consider programs as modules and define modular programs as sets of modules. The authors introduce ``input answer sets'', which is the key semantic object for communication between modules.

\cite{Cos2012ASP-modules} proposes to adopt ASP modules in order to simulate
(within reasonable complexity) possibility and necessity operators. Such operators (given the underlying modules) are meant to be usable in ASP programs, but possibly also programs written under other programming paradigms.

It can be seen that none of the above approach tackles modularization in
view of using ASP modules as standalone components in distributed systems. Therefore, our approach is a novelty in the landscape of the current literature.


\section{\Mus: Specification and Implementation Guidelines}
\label{muspecification}

In this section we provide an abstract definition of a \Mu, and some more specific indication of how such a component might be enacted and inserted into a distributed system, and how the inner ASP program might be structured.

\begin{definition}
Let $\Pi$ be an ASP program, and let $U = \mathit{Undef}(\Pi)$.
A \Mu\ based upon $\Pi$, denoted as \Mu($\Pi$), has the following specification:
\begin{itemize}
	\item Inner ASP program $\Pi$;
	\item Activation signal $A$ (optional), with $A \in \mathit{Undef}(\Pi)$;
	\item Stop signal $S$ (optional), with $S \in \mathit{Undef}(\Pi)$;
	\item Input set $\{I_1,\ldots,I_k\} \subseteq \mathit{Undef}(\Pi)$;
	\item Output set $\{O_1,\ldots,O_h\} \subseteq \mathit{Heads}(\Pi)$.
	\item Query result set $\{Q_1=v,\ldots,Q_r=v\}$ where $\{Q_1,\ldots,Q_r\}$ are
	queries\footnote{c.f. previous section for possible queries.} formulated over atoms occurring in $\mathit{Heads}(\Pi)$ and v = true/false.
\end{itemize}

\end{definition}

\noindent
The various elements listed above have the following meaning.

\begin{description}
\item 
Whenever the activation signal is specified, if $A$ is not true in $\Pi$,
then \Mu($\Pi$) is in a state of no-operation.
\item
Whenever the stop signal is specified, if $S$ becomes true in $\Pi$,
then \Mu($\Pi$) will go back into a state of no-operation.
\item
The input set is a set of atoms that, when some of them are added to $\Pi$, contribute to answer sets
computation. Each of such atom corresponds to an input/request received from the \Mu's surrounding environment.
\item
The output set is a set of atoms extracted from the answer sets of $\Pi$ plus the current input set.
Each of these atoms corresponds to an output/answer to be delivered into the \Mu's surrounding environment.
\item
The query result set is a set of truth values elicited from the answer sets of $\Pi$.
Each of these values corresponds to result of a query, to be delivered into the \Mu's surrounding environment.
\end{description}

In order to make it possible for \Mu($\Pi$) to operate dynamically, thus receiving inputs and delivering outputs and answers, a suitable \emph{shell} program must be defined, in any programming language able to be interfaced with an answer set solver.
Below we provide a schematic definition of such a shell program, to be used as a guideline for actual definition and implementation.

\begin{definition}
The shell responsible to manage an ASP microservice \Mu($\Pi$) can be specified
by the following pseudo-code. 

\meno
{\bf begin}\\
{\bf while} not activation then wait {\bf endwhile};\\
{\bf if} activation {\bf then} add atom $A$ to $\Pi$ as a fact to bring it into operation;\\
{\bf while} not stop do at frequency f\\
\tbm detect and annotate actual inputs $\{I_{j1},\ldots,I_{jr}\} \subseteq \{I_1,\ldots,I_k\}$;\\
\tbm add $\{I_{j1},\ldots,I_{jr}\}$ to $\Pi$ as facts;\\
\tbm obtain the answer sets $\{S_1,\ldots,S_n\}$ of (the augmented) $\Pi$;\\
\tbm elicit outputs $\{O_1,\ldots,O_v\} \subseteq \{O_1,\ldots,O_h\}$;\\
\tbm extract query results $\{Q_1,\ldots,Q_t\} \subseteq \{Q_1,\ldots,Q_r\}$;\\
\tbm deliver outputs and query results according to requests;\\
\tbm remove $\{I_{v_1},\ldots,I_{v_s}\} \subseteq \{I_{j_1},\ldots,I_{j_r}\}$ from $\Pi$ and remove relative annotations;\\
{\bf endwhile};\\
add atom $S$ to $\Pi$ as a fact  to bring it into no-operation;\\
{\bf end}.
\end{definition}

This shell program is able to activate and stop a \Mu, and to execute,
until possibly a stop signal arrives, a loop where the inputs are received from the external environment and delivered to $\Pi$, and outputs and query results are extracted from the answer sets of $\Pi$ (given the inputs) and delivered to the external environment. Precisely, each input will arrive from some other component, to which the output will have to be delivered. The shell program will rely upon an input-output table, where each input, expected output and related component will be annotated. Notice that at the end of each cycle some or all of the inputs will be removed from $\Pi$ and the relative annotations will be eliminated (removing all inputs determines a stateless component, while omitting to remove some of the inputs, forever of for some time interval, accounts to defining a stateful component). Input detection will occur at a certain frequency, suitable for each particular kind of component, environment, and application domain. Some of the inputs may come from sensors (and therefore they do not require any answer) and some of the outputs may go to actuators. This is also annotated in the input-output table.
Notice also that the parts concerning the activation and stopping of the \Mu\ (first and second line after the {\bf begin}, and last line before the {\bf end}) will be omitted if the component is running forever rather being first activated and then stopped. 

Many practical aspects remain to be defined in order to obtain an
implementation. For instance, if a \Mu\ is to be situated within 
a multi-agent system, input-output-query exchange might happen via
FIPA (cf. \url{http://www.fipa.org} for language specification, syntax and semantics), a widely used ACL (Agent Communication Language). The shell
program can be made FIPA-compliant (i.e., able to exchange and understand 
FIPA messages) for instance by importing the freely available JADE library (cf. \url{https://jade.tilab.com/} where references to several related publications can also be found). The JADE library is an advance middleware that 
offers many functionalities to \as agentify'' imperative or object-oriented programs. In fact it provides: the agent abstraction (i.e., a given program, when running, is seen by the external environment as an agent); the ability of peer to peer inter-agent FIPA asynchronous message-passing; a yellow pages service supporting
subscription of agents and a discovery mechanism, and many other facilities to support the development of distributed systems. 

So for instance, an input can be sent to a \Mu\ via a FIPA \emph{request} message with the input as argument, to be
interpreted on the \Mu's side as a request to reply with a \emph{confirm}
message containing the corresponding output. A query can be sent to the \Mu\ 
via a \emph{query-if} message whose answer will be a \emph{confirm}, conveying
the truth value of the query. Notice that, to avoid ambiguities, the FIPA syntax provides the facility to identify each message via a certain arbitrary identifier, so that the answer message can indicate that it is `in-reply-to' to that identifier.

The JADE yellow pages services might be exploited by \Mus\ which 
would want to register as agents with a name and a role, and then communicate with each other in an asynchronous way. Or, since most MAS offer such a mediator
service, \Mus\ might enroll in any known agent community. Finally, they might communicate peer-to-peer with other agents that they are aware of, or that they locate via the mediator.

Let us now consider how to structure `core' program
$\Pi$, on which a microservice \Mu($\Pi$) is based.
First, activation and stopping of a module can be simply obtained
by a couple of constraints, that make the program inconsistent
(in no-operation state) if either activation $A$ has not arrived, or stopping
signal $S$ has been issued: 

\st $
\begin{array}{l}
\prol not\ A.\ \ \ \ \ \%\ \mathit{module\ activation}\\
\prol S.\ \ \ \ \ \ \ \ \ \ \ \ \%\ \mathit{module\ stop}
\end{array}$

Then, when the module has been activated, upon arrival of new inputs, the 
inner program $\Pi$ will in general `produce' (admit) answer sets. If the answer set
is unique then the outputs can be univocally identified. Otherwise the shell,
in the `elicit outputs' part, will have to adopt some kind of policy
(e.g., preferences, utilities, costs or other) to select which answer set to
consider. The queries, being by definition defined upon the whole set
of answer sets, will always return an univocal result. In case, given the present input,
$\Pi$ should be inconsistent, then the output will consist in a failure signal
(e.g., in the FIFA ACL, there is the \emph{failure} primitive to be used in such cases).

\section{Case Study}
\label{casestudy}

The case study that we propose here is inspired to issues raised by 
applications related to autonomous vehicles.
Presently, machine learning mechanism have been defined to allow autonomous cars
to comply with traffic lights by detecting their color, so as to pass with green and stop with red similarly to traditional cars. Such mechanisms
must be trained, are prone to errors, and are potentially subject
to adversarial machine learning. 

In our view, physical traffic lights 
might in perspective disappear, to be substituted by monitoring
agents that would receive requests to pass from cars and consequently issue authorizations.
This either in routes dedicated to autonomous vehicles, or in the
(very reasonable) hypothesis to equip also `traditional' cars
with a device to interact with the monitoring agents.

Below we propose the sample design of the inner program concerning a \Mu\  
which implements the monitoring agent of a road intersection, taking the place
of a physical traffic light. 
In the example, the traffic light agent is called $tl$ and,
for the sake of simplicity, behaves like a `real' traffic light but
just takes the colors green ($g$ for short)
and red ($r$ for short). In fact, the yellow is no longer necessary as we assume
that the involved cars (each one equipped with its own driver agent)
will obey the directives. We have two lanes, one going north-south ($ns$ for short)
and the other one east-west ($ew$ for short), crossing at the traffic light.
If the traffic light is green in one direction it must be red in the 
other one, and vice versa. The traffic 
light is activated by a signal $active(t1)$, and never stopped unless there
is a fault, detected by the module itself by means of a sensor. 
A fault is supposed to have occurred whenever $fault\_tl$ is true,
i.e., it has been returned by the sensor. 

\st $
\begin{array}{l}
tln(t1).\ \ \ \%\mathit{Traffic-Light\ Identifier}\\
active(t1).\\
\\
\prol not\ active(t1).\ \ \ \%\ \mathit{Sensor\ Check\ activation}\\
\prol lane(L),fault\_tl(t1,L,T).\ \ \ \ \ \%\ \mathit{Sensor\ Check\ Possible\ Fault}
\end{array}$

\medskip
Each car, say here $c1$, $c2$, $c3$, $c4$ and $c5$, wants to go, but it is allowed to proceed only if it gets the green traffic light.
Otherwise, it remains dummy. We assume that all cars behave in the same way. 
Each one issues a request of format $car(C),want\_go(C,t1,L,T)$ where $L$ is the lane, with possible values $ns$ for north-south and $ew$ for east-west; $T$ is the time of the request. Requests by various cars may for example give rise to the addition of the following facts to the \Mu's program.

\st $
\begin{array}{l}
\% \mathit{INPUT:\ CARS}\\
car(c1).\\
car(c2).\\
car(c3).\\
car(c4).\\
car(c5).
\end{array} $

\st $
\begin{array}{l}
\% \mathit{INPUT:\ REQUESTS}\\
want\_go(c1,t1,ns,2).\\
want\_go(c2,t1,ns,2).\\
want\_go(c3,t1,ew,2).\\
want\_go(c4,t1,ns,4).\\
want\_go(c5,t1,ew,4).
\end{array} $

\medskip
The following facts and rules define the lanes, and specify that
this monitoring agent has a lookahead of five time instants:
after that, it will have to be re-run.

\st $
\begin{array}{l}
lane(ns).\\
lane(ew).\\
time(1..5).\\
next(Y,X) \prol time(X), time(Y), Y = X + 1.\\
\end{array}
$

\medskip
Rules below define the color that the traffic light takes in a very standard way as transitions from green to red and vice versa, where the initial color is green. In reality, such a monitoring agent can employ a much more sophisticated protocol such as for instance the Contract Net Protocol (CNP). If adopting CNP, the agent might grant priority to particular kinds of vehicles, e.g., police cars, ambulances, cars transporting a disabled person, etc. More generally, any policy to grant passage according to criteria could be implemented. 

\st $
\begin{array}{l}
tl(r,TL,L1,T1)\prol\\ 
\tbl time(T),lane(L1),lane(L2),tln(TL),L1!=L2,next(T1,T),\\
\tbl tl(g,TL,L1,T),tl(r,TL,L2,T).\\
tl(g,TL,L1,T1)\prol\\ \tbl time(T),lane(L1),lane(L2),tln(TL),L1!=L2,next(T1,T),\\
\tbl tl(r,TL,L1,T),tl(g,TL,L2,T).\\
tl(g,TL,ns,1)\prol tln(TL).\\
tl(r,TL,ew,T)\prol tln(TL),time(T),tl(g,TL,ns,T).
\end{array}$

\medskip
In our case the implemented protocol is fair, as cars that cannot go now because it is red on their lane
will be deferred to the next time instant (by delaying their request), when the color will be green
(output in format $go(Car,t1,Lane,Time)$).

\st $
\begin{array}{l}
go(C,TL,L,T)\prol\\
\tbl time(T),car(C),tln(TL),lane(L),want_go(C,TL,L,T),tl(g,TL,L,T).\\
wait(C,TL,L,T)\prol\\
\tbl time(T),car(C),tln(TL),lane(L),want_go(C,TL,L,T),tl(r,TL,L,T).\\
want\_go(C,TL,L,T1)\prol car(C),tln(TL),lane(L),wait(C,TL,L,T),next(T1,T).\\
\\
\prol time(T),car(C),tln(TL),lane(L),go(C,TL,L,T),tl(r,TL,L,T).
\end{array} $

Clearly, this program can be `cloned' (\emph{mutatis mutandis}) to manage any other traffic lights. For the reader's convenience, this program is standalone and can be run exactly as it is to check its results.

\section{System's Semantics }
\label{extendedMCS}
\label{semantics}

The semantics of a single\Mu\ is fully specified by: (i) the answer sets of the inner ASP program; (ii) the policy employed in its shell to select one single answer set; (iii) the set of queries that the shell performs over the entire set of answer sets, whose meaning is formally specified in \cite{CostantiniF16,Costantini19}. We aim however to provide a semantics for the overall distributed system composed of heterogeneous microservices, in order to provide a firm ground and a guideline for implementation. 

To do so, we resort to Multi-Context Systems (MCSs), that are a well-established paradigm in Artificial Intelligence and Knowledge Representation, aimed to model information exchange among heterogeneous sources \cite{BrewkaE07,BrewkaEF11,BrewkaEP14}.
However, with some abuse of notation (and some slight loss of generality) we adapt and readjust the definitions to fit into our framework. To represent the heterogeneity of sources, each component in a Multi-context system, called `context',is supposed to be based on its own \emph{logic}, defined in a very general way~\cite{BrewkaEF11}. In particular, a logic is defined out of the following.
\begin{itemize}
\item  A set $F$ of possible \emph{formulas} (or $\mathit{KB}$-elements)
under some signature;
\item A set $KB$ of knowledge bases built out of elements of  $F$. in our framework, $KB$ can also be a program in some programming language;
\item A function $\ACC$, where $\ACC(kb,s)$ means that $s$ is an acceptable set of consequences of knowledge base $kb \in KB$, i.e., $s \subseteq Cn$, where $Cn$ is the set of all possible consequences that can be drawn from $kb$. We assume here that $\ACC$\ produces a unique set of consequences. In case of a program written in a non-logical programming language, such set can be the set of legal outputs given some input (added to $kb$), that will be a subset of all possible outputs $Cn$; for logical components, it will be (one of) the $kb$ model(s). For instance, as we have seen the shell of a \Mu\ will produce as consequences the elements occurring in the answer set selected according to some policy, along with query results (where queries deliver conclusions drawn in general from the whole set of answer sets)
\end{itemize}

A (Managed) \emph{multi-context system} (MCS) \(M = \{C_1,\ldots,C_{r}\}\) is a set of $r=|M|$ \emph{contexts}, each of them of the form $C_i = \langle c_i,L_i,kb_i,br_i\rangle$,
where:
\begin{itemize}
    \item $c_i$ is the context \emph{name} (unique for each context; if a specific name is omitted, index $i$ can act as a name). In \cite{CabalarCGF19} a context's name, in the definition of a bridge rule, can be a term called \emph{context designator} denoting the \emph{kind} of context that should deliver that data item (for instance, \emph{mycardiologist(c)}, \emph{customercare(c)}, \emph{helpdesk(h)}, etc.). 
    \item $L_i$ is a logic.
    \item $kb_i \in KB$ is a knowledge base.
    \item $br_i$ is the set of \emph{bridge rules} this context is equipped with. 
\end{itemize}  
 
 Contexts in an MCS are meant to be heterogeneous distributed components, that exchange data. In fact,
bridge rules are the key construct of MCSs, as it describes in a uniform way the communication/data exchange patterns between contexts.
Each bridge rule $\rho \in br_i$ has the form
\begin{eqnarray}
op_i(s) \ar (c_1: p_1), \ldots, (c_j : p_j)
\end{eqnarray}
where the left-hand side $s$ is called the \emph{head}, and 
the right-hand side is called the \emph{body},
and the comma stands for conjunction. The meaning is that, each data item $p_j$ is supposed to come from a context $c_j$\footnote{In the original formulation, there are additional literals $\no (c_1: p_{j+1}), \ldots, \no (c_j : p_n)$ meaning that in order for the bridge rule to be applied, the $p_{j+1}\ldots p_n$ must be false in the relative contexts. We disregard this part, as non-logical components cannot use logical negation. There is no loss of generality however, as each of the $p_1,\ldots,p_j$ can state a negative fact.}. Whenever all the $c_1,\ldots,c_j$ have delivered their data item to the \emph{destination context} $c_i$, the rule becomes \emph{applicable}. When the rule is actually applied, its conclusion $s$, once elaborated by operator $Op$, will constitute an input to $c_i$. Operator $op_i$ can perform any elaboration on \as raw'' input $s$, such as format conversion, filtering, elaboration via ontologies, etc. Its operation is specified via a \emph{management function} $mng_i$, which is therefore crucial for knowledge incorporation from external sources. For simplicity, here we assume $mng_i$ to be monotonic (i.e., to produce from $s$ one or more data items) . Therefore, we can extend the previous definition of a context as $C_i = \langle{c_i,L_i,kb_i,br_i,mng_i}\rangle$. 
A bridge rule is \emph{applicable} if its body its true, in operational terms if all the data items listed in its body have arrived. In case context designators are employed, prior to checking a bridge rule for applicability, such terms must be substituted by actual context names. For \Mus, this task will be performed by the shell, that must then be endowed with a list of contexts of each type. 

A \emph{data state} (or \emph{belief state}) $\vec{S}$ of an MCS~$M$
is a tuple $\vec{S} = (S_1,\ldots, S_{r})$ such that for $1 \leq i \leq {r}$, $S_i \subseteq Cn_{i}$. A data state can be seen as a view of the distributed system by an external \as observer''.
$\mathit{app}(\vec{S})$ is the set composed of the head of those bridge rules which are \emph{applicable} in~$\vec{S}$. This means, in logical terms, that their body is true w.r.t. $\vec{S}$. In practical terms, we may say that a bridge rule $\rho$ associated to context $c_i$ is applicable in $\vec{S}$ if all the data mentioned in the body of the bridge rule can be delivered to the destination context. This is the case whenever they are available in the contexts of origin, i.e.,  they occur in the present respective data state items in $\vec{S}$. In the original formulation of MCS, all applicable bridge rules are automatically applied, and their results, after the elaboration by the management function, are added to the destination context's knowledge base, that therefore grows via bridge-rule application. 

Starting from a certain specific data state, some bridge rules will be applicable and therefore they will be applied. This will enhance the knowledge base in some of the contexts, thus determining (in these contexts) a new set of acceptable consequence, and therefore a new overall data state. In the new state other bridge rules will be applicable, and so on, until a \as stable'' state, called \emph{Equilibrium}, will be reached. Technically, $\vec{S}$ is an \emph{equilibrium} for an mMCS $M$ iff, for $1 \leq i \leq |M|$,
\begin{eqnarray}
S_i = ACC_i(mng_i(\mathit{app}(\vec{S}),kb_i)) 
\end{eqnarray}which states that each element of the equilibrium is an acceptable set of consequences after the application of every applicable bridge rule, whose result has been incorporated into the context's knowledge base via the management function. in \cite{CabalarCGF19} it is proved that, in the kind of MCS that we have just described, an equilibrium will be reached in a finite number of steps. Notice however that this definition assumes the system to be isolated from any outside influence, and that an equilibrium, one reached, will last forever. Instead, in real systems there will be interactions with an external environment, and so equilibria may change over time. Moreover, each context is not necessarily a passive receiver of data sent by others.

To take these aspects into account, \cite{CabalarCGF19} proposes some extensions to the original formulation, among which the following, that are relevant in the present setting.
\begin{itemize}
    \item 
    It is noticed that contexts knowledge bases 
    can evolve in time, not only due to bridge-rule
    application. In fact, contexts receive sensor inputs (passively or in consequence to active observation), or can be affected by user's modification (e.g.,a context encompasses a relational database that can be modified by a user). So, each context $c_i$ will have an associated \emph{Update Operator} $\mathit{{\cal U}_i}$ (that can actually consist in a tuple of operators performing each one a different kind of update). Updates and bridge rules both affect contexts' knowledge base over time. So (assuming an underlying discrete model of time) we will be able to consider, when necessary, $c_i[T]$ meaning context $c_i$ at time $T$, with its knowledge base $kb_i[T]$; consequently we will have an evolution over time of contexts. Therefore, we will have a definition (not reported here) of \emph{Timed Equilibria}. Notice only that a timed equilibrium can be reached at time $T+1$ only if the actual elapsed time between $T$ and $T+1$ is sufficient for the system to \as stabilize'' by means of bridge-rules application on the updated knowledge bases.
\item  Mandatory bridge-rule application constitutes a limitation: in fact, this forces contexts to accept inputs unconditionally, and this be often inappropriate. Consider for instance a context representing a family doctor: the context may accept non-urgent patient's requests for appointments or consultation only within a certain time windows. So, \cite{CabalarCGF19} introduces conditional bridge-rule application, formalized via a  \emph{timed triggering function},\,
\(\mathit{tr_i}\),
which specifies which applicable bridge rules are triggered (i.e., they are practically applicable) at time $T$. It does so either based on certain pre-defined conditions, or by performing some reasoning over the present knowledge base contents. Therefore, the implementation of \(\mathit{tr_i[T]}\) my require an auxiliary piece of program, that in a \Mu's shell will presumably be a logic program.
\end{itemize}

So, considering contexts which are \Mus, in order to fit in the vision of the overall system as an MCS, their shell must be empowered as follows.

\begin{itemize}
    \item Include the bridge rules associated to a\ \Mu, and the definition/implementation of the triggering function.
\item Include a facility to resolve the context designators, so as to check for applicability a triggered bridge rule after substituting context designators occurring therein with actual contexts' names.
\item Include the definition of the specific management function, so as to be able to apply it on bridge-rules' results.
\end{itemize}

In the case study of previous section, each traffic light should be equipped with a bridge rule that, by means of a suitable context designator (say, \emph{anycar(c)}) collects the cars' requests. Symmetrically, cars should be equipped with a bridge rule to collect the permission to go by the traffic light (the nearest one, whose identifier should replace a context designator of the form, e.g., \emph{a\_traffic\_light(t)}. The  triggering function may allow cars to enable receiving traffic-light communications only when this is deemed appropriate. 

Context designators are therefore useful to write general bridge rules to be then customized to the particular situation at hand. They also allow to devise a system where components do not know or are aware of each other in advance, and where components can possibly join/leave the system at any time.
A suitable middleware should be realized to allow component's shells to instantiate bridge rules. Concerning our case study, that concerns an infrastructure for autonomous vehicles, both cars and traffic lights might for instance broadcast their name and geo-localization. In this way, cars might locate the traffic light of interest, and traffic lights might become aware of nearby cars that might send them a request.

\section{Concluding Remarks}  
\label{conclusions}

We have proposed a methodology for developing microservices in Answer Set Programming, by means of the creation of a particular kind of components, that can be activated/stopped, can receive external requests and can deliver answers. We have provided a definition of a \Mu\ and explained how it might be implemented, and we have outlined a programming methodology.
We have also outlined a possible uniform semantics to specify an heterogeneous system in which \Mus\ can be situated. This is an absolute novelty for microservices in general, as no attempt has ever been made to provide such a uniform model for an overall system. The proposed semantics can constitute the ground for principled implementations. Overall, this work can be considered as a creative combination of existing technologies, in view of entirely new fields of application of answer set programming and logic programming in general.

Important application fields for \Mus\ are Cloud computing and IoT. We consider particularly important the various kinds of robotic applications and the underlying infrastructural aspects, as shown in the case study related to autonomous vehicles.

Future work includes refining the programming methodology on the one hand, 
devising a prototype implementation and experimenting the integration in realistic 
environment: we plan in particular to experiment the integration into DALI Multi-Agent-Systems, where DALI \cite{jelia02,jelia04,postdalt06,daliDOI} is the agent-oriented programming language
defined and implemented by our research group, that has been made compatible
with the Docker technology. This will allow us to experiment \Mus\ as components in the various applications where DALI is being applied, including cognitive robotic architectures.



\end{document}